\documentclass{article} % For LaTeX2e
\usepackage[accepted]{icml2021}

\usepackage{hyperref}
\usepackage{url}
\usepackage{xcolor}
\usepackage{multirow}
\usepackage{graphicx}
\usepackage{amsfonts}
\usepackage{amsmath}
\usepackage{bbm}
\usepackage{wrapfig}
\usepackage{booktabs}
\usepackage{subfig}

\newcommand{\method}[0]{Learned Belief Search}
\newcommand{\accmethod}[0]{LBS}

\icmltitlerunning{\method: Efficiently Improving Policies in Partially Observable Settings}

\begin{document}

\twocolumn[
\icmltitle{\method: \\
Efficiently Improving Policies in Partially Observable Settings}

\begin{icmlauthorlist}
\icmlauthor{Hengyuan Hu*}{fb}
\icmlauthor{Adam Lerer*}{fb}
\icmlauthor{Noam Brown}{fb}
\icmlauthor{Jakob Foerster}{fb}

\end{icmlauthorlist}

\icmlaffiliation{fb}{Facebook AI Research}

\icmlcorrespondingauthor{Hengyuan Hu}{hengyuan@fb.com}

\vskip 0.3in
]

\printAffiliationsAndNotice{\icmlEqualContribution} % otherwise use the standard text.

% \maketitle
\begin{abstract}
Search is an important tool for computing effective policies in single- and multi-agent environments, and has been crucial for achieving superhuman performance in several benchmark fully and partially observable games. However, one major limitation of prior search approaches for partially observable environments is that the computational cost scales poorly with the amount of hidden information. In this paper we present \emph{Learned Belief Search} (LBS), a computationally efficient search procedure for partially observable environments. Rather than maintaining an exact belief distribution, \accmethod~uses an approximate auto-regressive counterfactual belief that is learned as a supervised task. In multi-agent settings, \accmethod~uses a novel public-private model architecture for underlying policies in order to efficiently evaluate these policies during rollouts. In the benchmark domain of Hanabi, \accmethod~can obtain $55\% \sim 91\%$ of the benefit of exact search while reducing compute requirements by $35.8 \times \sim 4.6 \times$, allowing it to scale to larger settings that were inaccessible to previous search methods.
\end{abstract}
\section{Introduction}
\label{sec:intro}
% My typical: "let's move beyond 2-player-zero-sum rant. Because, why not? Also: Search is great
% Search for the win
Search has been a vital component for achieving superhuman performance in a number of hard benchmark problems in AI, including Go~\citep{silver2016mastering,silver2017mastering,silver2018general}, Chess~\citep{campbell2002deep}, Poker~\citep{moravvcik2017deepstack,brown2017superhuman,brown2019superhuman}, and, more recently, self-play Hanabi~\citep{lerer2020improving}. Beyond achieving impressive results, the work on Hanabi and Poker are some of the few examples of search being applied in large partially observable settings. In contrast, the work on belief-space planning typically assumes a small belief space, since these methods scale poorly with the size of the belief space.

% LBS = Scalable SPARTA
Inspired by the recent success of the SPARTA search technique in Hanabi~\citep{lerer2020improving}, we propose \emph{\method} (\accmethod) a simpler and more scalable approach for policy improvement in partially observable settings, applicable whenever
a model of the environment and the policies of any other agents are available at test time.
% LBS = Learned belief model 
Like SPARTA, the key idea is to obtain Monte Carlo estimates of the expected return for every possible action in a given action-observation history (AOH) by sampling from a belief distribution over possible states of the environment.
However, \accmethod~addresses one of the key limitations of SPARTA. Rather than requiring a sufficiently small belief space, in which we can compute and sample from an exact belief distribution, \accmethod{} samples from a learned, auto-regressive belief model which is trained via supervised learning (SL). 
The auto-regressive parameterization of the probabilities allows \accmethod{} to be scaled to high-dimensional state spaces, whenever these are composed as a sequence of features. 

%Bootstrapping rather than MC
Another efficiency improvement over SPARTA is replacing the full rollouts with partial rollouts that bootstrap from a value function after a specific number of steps. While in general this value function can be trained via SL in a separate training process, this is not necessary when the blueprint (BP) was trained via RL. In these cases the RL training typically produces both a policy and an approximate value function (either for variance reduction or for value-based learning). In particular, it is common practice to train \emph{centralized} value functions, which capture the required dependency on the sampled state even when this state cannot be observed by the agent during test time.

% Single Player policy improvement != POMDP
While \accmethod{} is a very general search method for Partially Observable Markov Decision Processes (POMDPs), our application is focused on single-agent policy improvement in Decentralized POMDPs (Dec-POMDPs) (in our specific case, Hanabi). 
One additional challenge of single-agent policy improvement in Dec-POMDPs is that, unlike standard POMDPs, the Markov state $s$ of the environment is no longer sufficient for estimating the future expected return for a given AOH of the searching player. Instead, since the other players' policies also depend on their entire action-observation histories, \emph{e.g.}, via Recurrent-Neural Networks (RNNs), only the union of Markov state $s$ and all AOHs is sufficient. 

%Revealing information flow. 
This in general makes it challenging to apply \accmethod, since it would require sampling entire AOHs, rather than states. However, in many Dec-POMDPs, including Hanabi, information can be split between the common-knowledge (CK) trajectory and private observations. Furthermore, commonly information is `revealing', such that there is a mapping from the most recent private observation and the CK trajectory to the AOH for each given player. In these settings it is sufficient to track a belief over the union of private observations, rather than over trajectories, which was also exploited in SPARTA. 
%SPARTA%LBS exploits this
We adapt \accmethod{} to this setting with a novel public-RNN architecture which  makes replaying games from the beginning, as was done in SPARTA, unnecessary.

When applied to the benchmark problem of two player Hanabi self-play, \accmethod{} obtains $55\%\sim91\%$ of the benefit of exact search while reducing compute requirements by $35.8\times \sim 4.6 \times$ depending on the trade-offs between speed and performance. 
We also successfully apply \accmethod{} to a six- and seven-card versions of Hanabi, where calculating the exact belief distribution would be prohibitively expensive.

\section{Related Work}
\label{sec:related_work}

% The 1-ply policy improvement search we use is similar prior work by \citet{lerer2020improving}, but we use a learned approximate belief and a bootstrapped value function that allows efficient scaling to larger games.

\subsection{Belief Modeling \& Planning in POMDPs}

% https://scholar.google.com/scholar?hl=en&as_sdt=0%2C5&q=belief+planning+pomdp&btnG

Deep RL on POMDPs typically circumvents explicit belief modeling by using a policy architecture such as an LSTM that can condition its action on its history, allowing it to implicitly operate on belief states \citep{hausknecht2015deep}. `Blueprint' policies used in this (and prior) work take that approach, but this approach does not permit search since search requires explicitly sampling from beliefs in order to perform rollouts.

There has been extensive prior work on learning and planning in POMDPs. Since solving for optimal policies in POMDPs is intractable for all but the smallest problems, most work focuses on approximate solutions, including offline methods to compute approximate policies as well as approximate search algorithms, although these are still typically restricted to small grid-world environments  \citep{ross2008online}.

One closely related approach is the Rollout algorithm \citep{bertsekas1999rollout}, which given an initial policy, computes Monte Carlo rollouts of the belief-space MDP assuming that this policy is played going forward, and plays the action with the highest expected value. In the POMDP setting, rollouts occur in the MDP induced by the belief states\footnote{SPARTA's single agent search uses a similar strategy in the DEC-POMDP setting, but samples states from the beliefs rather than doing rollouts directly in belief space.}.

There has been some prior work on search in large POMDPs. \citet{silver2010monte} propose a method for performing Monte Carlo Tree Search in large POMDPs like Battleship and partially-observable PacMan. Instead of maintaining exact beliefs, they approximate beliefs using a particle filter with Monte Carlo updates. \citet{roy2005finding} attempt to scale to large belief spaces by learning a compressed representation of the beliefs and then performing Bayesian updates in this space. 

Most recently MuZero combines RL and MCTS with a learned implicit model of the environment \citep{schrittwieser2019mastering}. Since recurrent models can implicitly operate on belief states in partially-observed environments, MuZero in effect performs search with implicit learned beliefs as well as a learned environment model.

%https://papers.nips.cc/paper/4031-monte-carlo-planning-in-large-pomdps.pdf

\subsection{Games \& Hanabi}

Search has been responsible for many breakthroughs on benchmark games. Most of these successes were achieved in fully observable games such as Backgammon \citep{tesauro1994td}, Chess \citep{campbell2002deep} and Go \citep{silver2016mastering,silver2017mastering,silver2018general}. More recently, belief-based search techniques have been scaled to large games, leading to breakthroughs in poker  \citep{moravvcik2017deepstack,brown2017superhuman,brown2019superhuman} and the cooperative game Hanabi \citep{lerer2020improving}, as well as large improvements in Bridge \citep{tian2020joint}.

There has been a growing body of work developing agents in the card game Hanabi. While early hobbyist agents codified human conventions with some search \citep{quuxplusone,FireFlower}, more recent work has focused on Hanabi as a challenge problem for learning in cooperative partially-observed games~\citep{bard2019hanabi}. In the self-play setting (two copies of the agent playing together), the Bayesian Action Decoder (BAD) was the first learning agent~\citep{foerster2019bayesian}, which was improved upon by the Simplified Action Decoder (SAD)~\citep{Hu2020Simplified}. The state-of-the-art in Hanabi self-play is achieved by the SPARTA Monte Carlo search algorithm applied to the SAD blueprint policy \citep{lerer2020improving}.

There has been recent work on ad-hoc team play and zero-shot coordination, in which agents are paired with unknown partners \citep{canaan2019diverse,walton2017evaluating,hu2020other}.
\section{Setting and Background}
\label{sec:background}

In this paper we consider a Dec-POMDP, in which $N$ agents each take actions $a_t^i$ at each timestep, after which the state $s_t$ updates to $s_{t+1}$ based on the conditional probability $P(s_{t+1} | s_t, \mathbf{a}_t)$ and the agents receive a (joint) reward $r(s_t, \mathbf{a_t})$, where $\mathbf{a}_t$ is the joint action of all agents.

Since the environment is partially observed each agent only obtains the observation $o_t^i = Z(s_t, i)$ from a deterministic observation function $Z$. We denote the environment trajectory as $\tau_t =\{s_0, \mathbf{a}_0, ... s_t,  \mathbf{a}_t\}$ and the action-observation history (AOH) of agent $i$ as $\tau^i_t =\{o_0^i,  {a}^i_0, ...,  o_t^i,  {a}^i_t\}$. The total forward looking return from time $t$ for a trajectory $\tau$ is $R^{t}(\tau)=\sum_{t'\geq t} \gamma^{t'-t} r(s_t, \mathbf{a_t})$, where $\gamma$ is an optional discount factor. 
Each agent chooses a policy $\pi^i(\tau^i)$ conditioned on its AOHs, with the goal that the joint policy $\pi=\{\pi^i\}$ maximises the total expected return $J_\pi = \mathbb{E}_{\tau \sim P(\tau|\pi)} R(\tau)$.

Starting from a common knowledge blueprint (BP), i.e. a predetermined policy that is known to all players, in order to perform \textit{search} in a partially observable setting, agents will need to maintain \textit{beliefs} about the state of the world given their observations. We define beliefs $B^i(\tau_t)=P((s_t, \{\tau_t^j\})|\tau_t^i)$, which is the probability distribution over states and AOHs, given player $i$'s private trajectory. Note that in Dec-POMDPs the beliefs must model other agents' AOHs as well as the current state, since in general the policies of other players condition on these AOHs. 

In general, the domain of $B^i$ (the \emph{range}) is extremely large, but in Dec-POMDPs with a limited amount of hidden information there is often a more compact representation. For example, in card games like Hanabi, the range consists of the unobserved cards in players' hands. SPARTA \citep{lerer2020improving} assumed that the domain was small enough to be explicitly enumerated. In our case, we assume that the beliefs are over \textit{private features} $f^i$ that can be encoded as a sequence of tokens from some vocabulary: $f^i = \prod_j f^i_j \in \mathcal{V}$. Furthermore, it simplifies the algorithm if, as is typically the case, we can factor out these private features from $\tau$ to produce a public trajectory $\tau^{pub}$; then each $\tau^i$ can be specified by the pair $(\tau^{pub}, f^i)$ and $\tau$ by $(\tau^{pub}, f^1, ..., f^N)$.

\accmethod{} is applicable to general POMDPs, which are a natural corner case of the Dec-POMDP formalism when we set the number of players to 1.

% consists of all possible trajectories, but in many games the elements of $B^i$ can be summarized by a small number of features; for example in Hanabi, given agent $i$s AOH $\tau_t^i$ each possible trajectory $\tau_t$ corresponds to a specification of the hidden cards in agent $i$s hand. In simple terms, all the hidden information in Hanabi consists of the player's unobserved cards. SPARTA \cite{lerer2020improving} leverages this fact to maintain explicit beliefs over this set of possible trajectories. We will instead represent the elements of this distribution as a sequence of $N$ features from some vocabulary $f=\mathcal{V}^N$

% \subsection{SPARTA} 
\section{Method}
\label{sec:method}

% \jakob{I think this should go into background? }

The starting point for our method is the single-agent search version of SPARTA \citep{lerer2020improving}: Given a set of BP policies, $\pi^i$, we estimate expected returns, $Q(a^i |\tau^i)$, by sampling possible trajectories $\tau$ from a belief conditioned on $\tau^i$:

\begin{equation}
       Q(a^i |\tau^i) = \mathbf{E}_{{\tau}\sim P(\tau | \tau^i )}Q(a^i | \tau ) 
\end{equation}

Here $ Q(a^i |\tau)$ is the expected return for playing action $a^i$ given the history $\tau$:

\begin{equation}    Q(a^i |\tau) = \mathbf{E}_{{\tau'} \sim P(\tau' | \tau, a^i )} R^t(\tau'),
\end{equation}
where $R^t(\tau')$ is the Monte Carlo forward looking return from time $t$. 

\begin{figure*}[t]
    % \vspace{-3.5em}
\centering
\includegraphics[width=0.95\linewidth]{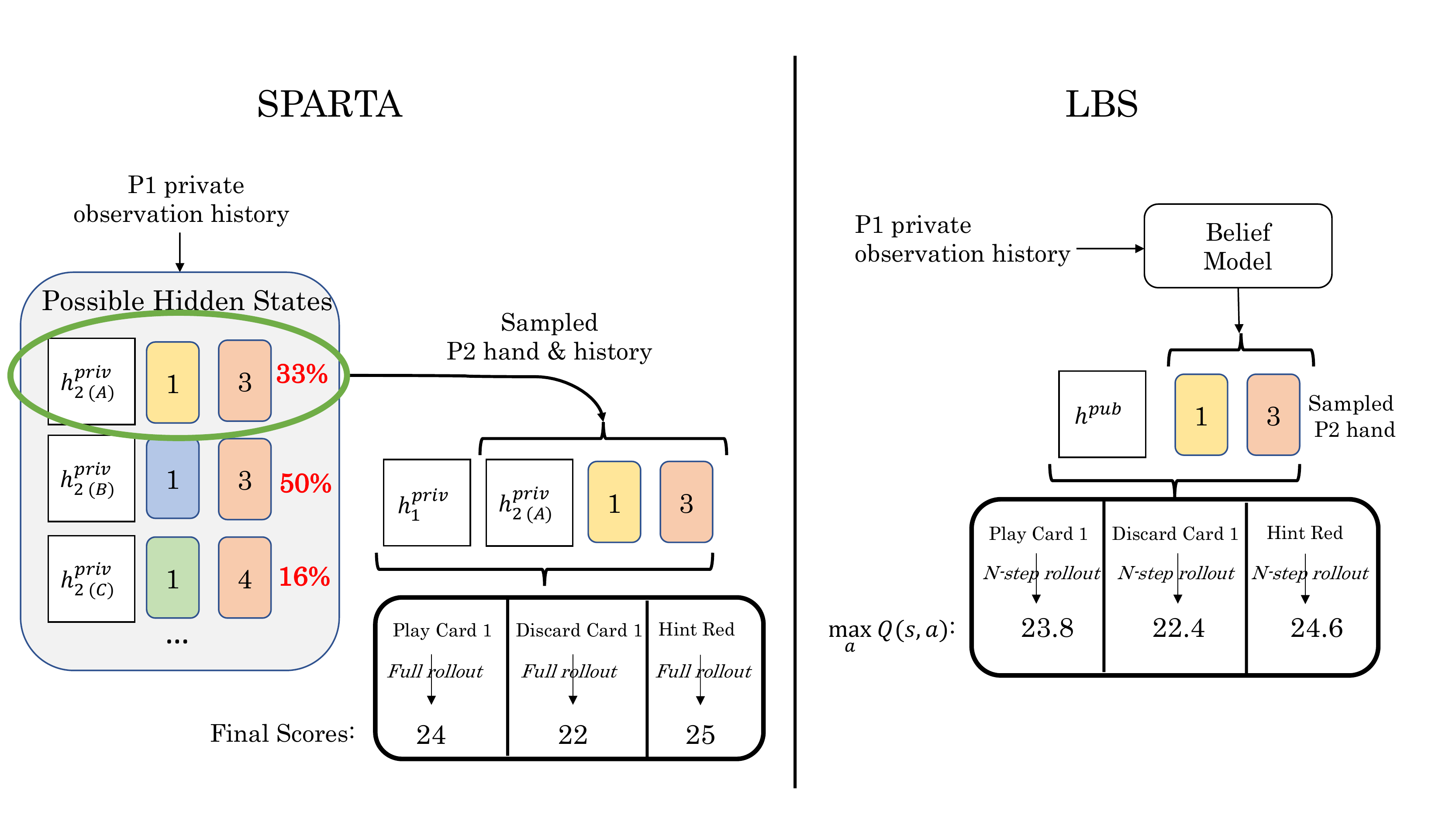}
% \vspace{-3mm}
\caption{\small{Comparison of SPARTA \citep{lerer2020improving} and LBS (ours). SPARTA maintains an explicit belief distribution with an accompanying AOH $h_2^{priv}$ for each belief state. LBS uses an auto-regressive belief model to sample states from the belief distribution, given the AOH. AOHs do not need to be maintained for each belief state in LBS since the model only relies on the public trajectory. Additionally, LBS uses an N-step rollout followed by a bootstrap value estimate.}}
\label{fig:speedy_search}
    % \vspace{-1.5em}
\end{figure*}

Whenever the argmax of $Q(a^i |\tau^i)$ exceeds the expected return of the BP, $Q(a_{BP} |\tau^i)$, by more than 
a threshold $\delta$, the search-player deviates from the BP and plays the argmax instead. 
For more details, please see Figure~\ref{fig:speedy_search} (LHS) and the original paper.

Even though this is single-agent search, in SPARTA this process is costly: First of all, the belief $P(\tau | \tau^i )$ is an exact counterfactual belief, which requires evaluating the BP at every timestep for every possible $\tau$ to identify which of these are consistent with the actions taken by the other agents. 
Secondly, to estimate $R^t(\tau')$ SPARTA plays full episodes (\emph{rollouts}) until the end of the game. 

Lastly, since in Dec-POMDPs policies condition on full AOHs (typically implemented via RNNs) the games have to be replayed from the beginning for each of the sampled $\tau$ to obtain the correct hidden state, $h(\tau^i(\tau))$, for all players $i$, for each of the sampled trajectories, $\tau$. 

\begin{figure*}[t]
    % \vspace{-3.5em}
\centering
\includegraphics[width=0.95\linewidth]{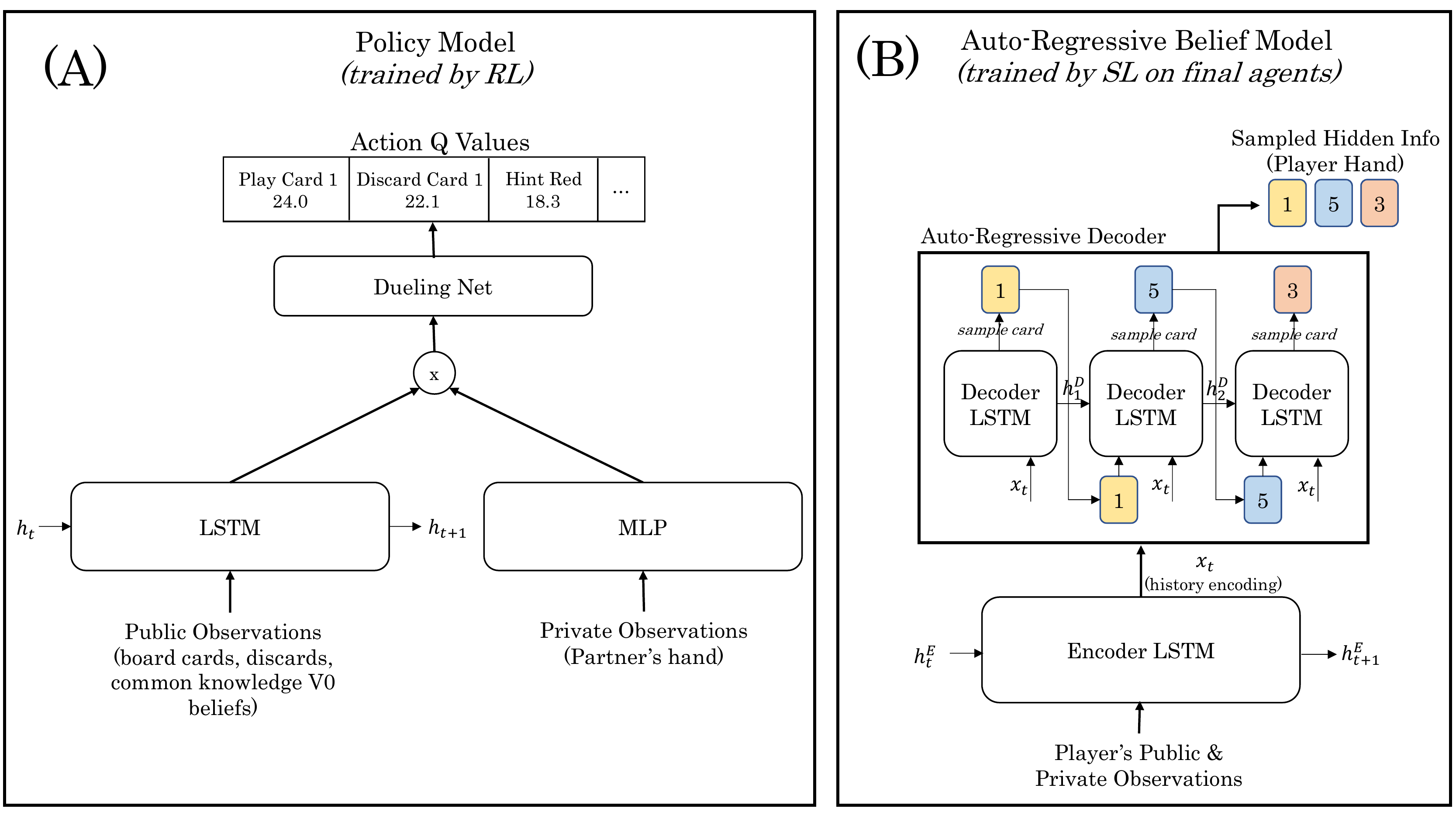}
% \vspace{-3mm}
\caption{\small{(\textbf{A}): Illustration of the public-LSTM network used for the BP policy. (\textbf{B}): The auto-regressive network for modeling beliefs .}}
    % \vspace{-1.5em}
\label{fig:network}
\end{figure*}

\method~addresses all of these issues: First of all, rather than using costly exact beliefs, we use supervised learning (SL) to train an auto-regressive belief model which predicts $P(\tau | \tau^i )$.
As described in Section~\ref{sec:background}, in our setting this reduces to predicting the \emph{private observations} for all other players $f^{-i}$, since the public trajectory is known. We use an auto-regressive model to encode the private observations as it can be decomposed to sequence of tokens $f^{-i} = \prod_{j} f_{j}^{-i}$. For scalability, as illustrated in Figure~\ref{fig:network} (RHS), the auto-regressive belief model is parameterized by a neural network with weights $\phi$:
\begin{equation}
P_{exact}(f^{-i} | \tau^i) \rightarrow  P_\phi(f^{-i} | \tau^i) = \prod_j P(f^{-i}_j | f^{-i}_{<j}, \tau^i) .
\end{equation}

Secondly, to avoid having to unroll episodes until the end of the game we use a learned value function to bootstrap the expected return after a predefined number, $N$, of steps. 
While this value function in general needs to be trained via SL, this is not necessary in our setting: Since our BPs are recurrent DQN agents that learn an approximate expected return via value-decomposition networks (VDN), we can directly use the BP to estimate the expected return, as illustrated in Figure~\ref{fig:speedy_search} (B) :
\begin{equation}
R^t(\tau') \simeq \sum^{t + N}_{t'=t} r_{t'} + \sum_i Q^i_{BP}(a^i_{BP} |f^i_{t+N}, \tau^{pub}_{t + N }) | \tau'.
\end{equation}
% The second term in the equation computes the joint Q values from the BP, which has been trained to estimate expected total return $R^{t+N+1}(\tau')$ under BP starting from $t+N+1$.
This is a fairly general insight, since RL training in POMDPs commonly involves centralized value functions that correctly capture the dependency of the expected return on the central state. 

Lastly, we address a challenge that is specific to Dec-POMDPs: To avoid having to re-unroll the policies for the other agents from the beginning of the game for each of the sampled $\tau$, \accmethod{} uses a specific RNN architecture.
Inspired by other public-private methods~\citep{foerster2019bayesian,kovavrik2019rethinking,horak2019solving}, we exploit that the public trajectory in combination with the current private observation contains the same information as the entire private trajectory and only feed the public information $\tau^{pub}$ into the RNN. The public hidden state $h(\tau^{pub}_t)$ is then combined with the private observation $f^i_t$ through a feedforward neural network, as illustrated in Figure~\ref{fig:network} (LHS).
\begin{equation}
    \pi(\tau^i_t) \rightarrow \pi(h(\tau^{pub}_t ), f^i_t)
\end{equation}

We note that whenever it is possible to factor out the public trajectory, this architecture can be used. If not, \accmethod{} can still be used, but instead for each sampled $f^i$ we would need to reconstruct the entire game from scratch to obtain the correct model state. 

We also point out that in this paper we only consider single player search where all others act according to the BP. Carrying out search for more than one player would not be theoretically sound because the trained belief model would no longer be accurate. Crucially, the learned belief model can only provide accurate implicit beliefs when the other players are playing according to the blueprint policy, because the belief model is trained before any agent conducts search. As a consequence, doing independent LBS for multiple players at the same time would lead to inaccurate beliefs. As we show later in the paper, it will lower the performance. Further details, including specific architectures, of our three innovations are included in Section \ref{sec:experiments}.

\section{Experimental Setup}
\label{sec:experiments}

We evaluate our methods in Hanabi, a partially observable fully cooperative multi-step board game.
In Hanabi, the deck consists of five different colors and five ranks from 1 to 5. For each color. There are three 1s, two 2s, 3s, 4s and one 5, totaling 50 cards. At beginning of a game, each player draw 5 cards. Each player cannot observe their own cards but instead can see other players' cards. Players take turn to move and the goal for the team is to play cards of each color from 1 to 5 in the correct order to get points. At each turn, the active player can either play a card, discard a card, or select a partner and reveal information about their cards. Playing a wrong card will cause the entire team to lose 1 life token. The game will terminate early if all 3 life tokens are exhausted and the team will get 0 point. To reveal information, the active player can either choose a color or a rank and {\it every} card of that color/rank will be revealed to the chosen player. The game starts with 8 information tokens. Each reveal action costs 1 token and each discard action regains 1 token. Players will draw a new card after play/discard action until the deck is finished. After the entire deck is drawn, each player has one last turn before the game ends.
For simplicity we focus on 2-player Hanabi for all our experiments and note that it is straightforward to extend our method to any number of players. 

%In particular, in Hanabi the size of the hidden information is reduced for 4-5 players, since in those cases players only have 4 cards in their hands. 

\subsection{Blueprint Training}
\label{sec:bp_training}
As explained in Section~\ref{sec:background} the public and private observations in Hanabi can be factorized into public and private features. We modify the open-source Hanabi Learning Environment (HLE)~\citep{bard2019hanabi} to implement this. 
Here, the only private observation is the partner's hand, while all other information is public.
There are many different options for implementing the public-RNN concept. We end up with the design shown in Figure~\ref{fig:network} (A). Specifically, the LSTM only takes public features as input while an additional MLP takes in the concatenation of private and public features. The outputs of the two streams are fused through element-wise multiplication before feeding into the dueling architecture~\citep{dueling-dqn} to produce Q values. We have also experimented with other designs such as using concatenation in place of the element-wise multiplication, or feeding only private observation to the MLP. Empirically we find that the design chosen performs the best, achieving the highest score in self-play.

We use the distributed training setting described in~\cite{obl}. We follow their hyper-parameters but replace the network with the one described above. The training jobs run with 40 CPU cores and 3 GPUs.

\subsection{Belief Training}
\label{sec:belief_training}

The belief model is trained to predict the player's own hand given their action observation history. An overview of the architecture is shown in the right panel of Fig~\ref{fig:network}. An encoder LSTM converts the sequence of observations to a context vector. 
The model then predicts its own hand in an auto-regressive fashion from oldest card to newest. The input at each step is the concatenation of the context vector and the embedding of the last predicted card. The model is trained end-to-end with maximum likelihood: \begin{equation}
\mathcal{L}(c_{1:n} | \tau) = -\sum_{i=1}^{n} \log p(c_i | \tau, c_{1:i-1}),
\label{eq:belief-loss}
\end{equation}
where $n$ is the number of cards in hand and $c_i$ is the value of the $i$-th card.

We use a setup similar to that of reinforcement learning to train the belief model instead of a more traditional way of creating a fixed train, test and validation set. We use a trained policy and a large number of parallel Hanabi simulators to continuously collect trajectories and write them into a replay buffer. In parallel we sample from the replay buffer to train the belief model using a supervised loss. This helps us easily avoid over-fitting without manually tuning hyper-parameters and regularization. The RL policy used to generate data is fixed during the entire process. It takes 20 CPU cores and 2 GPUs to train the belief models.

\subsection{\accmethod{} Implementation Details} 

\method~is straightforward once we have trained a BP and a belief model. 
The search player samples hands from the belief model and filters out the ones that are inconsistent with current game status based on their private observation. In the extremely rare case where the belief model fails to produce a sufficient number of legal hands, it reverts back to the BP. To understand the impact of various design parameters, we experimented with both playing out all trajectories until the end of the game as well as rolling out for a fixed number of steps and using a bootstrapped value estimate at the final state. Similar to SPARTA, the search actor only deviates from the BP if the expected value of the action chosen by search is $\delta=0.05$ higher than that of the BP and a UCB-like pruning method is used to reduce the number of samples required. All search experiments are executed using 5 CPU cores, 1 GPU and 64GB of memory.
\section{Results}
\label{sec:results}
In this section, we first show the performance and run time of our method compared with blueprint baseline and exact search method, SPARTA. We then study the quality of the learned belief by comparing it to two analytical benchmarks. Finally, we study trade-offs between training and testing under a fixed budget.

\subsection{Performance} 
\begin{figure}[t]
 \centering
  \includegraphics[width=0.42\textwidth]{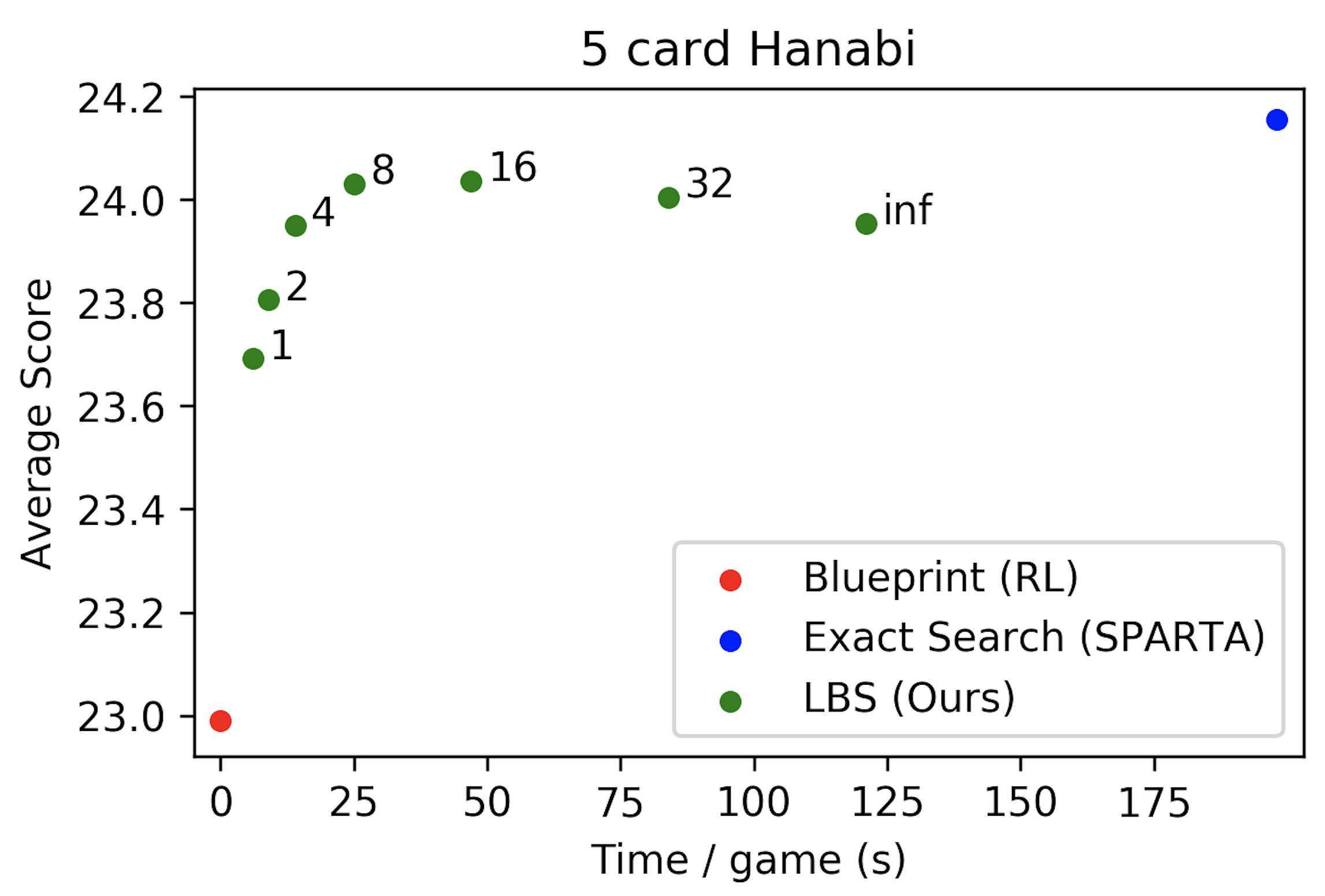}
  \includegraphics[width=0.45\textwidth]{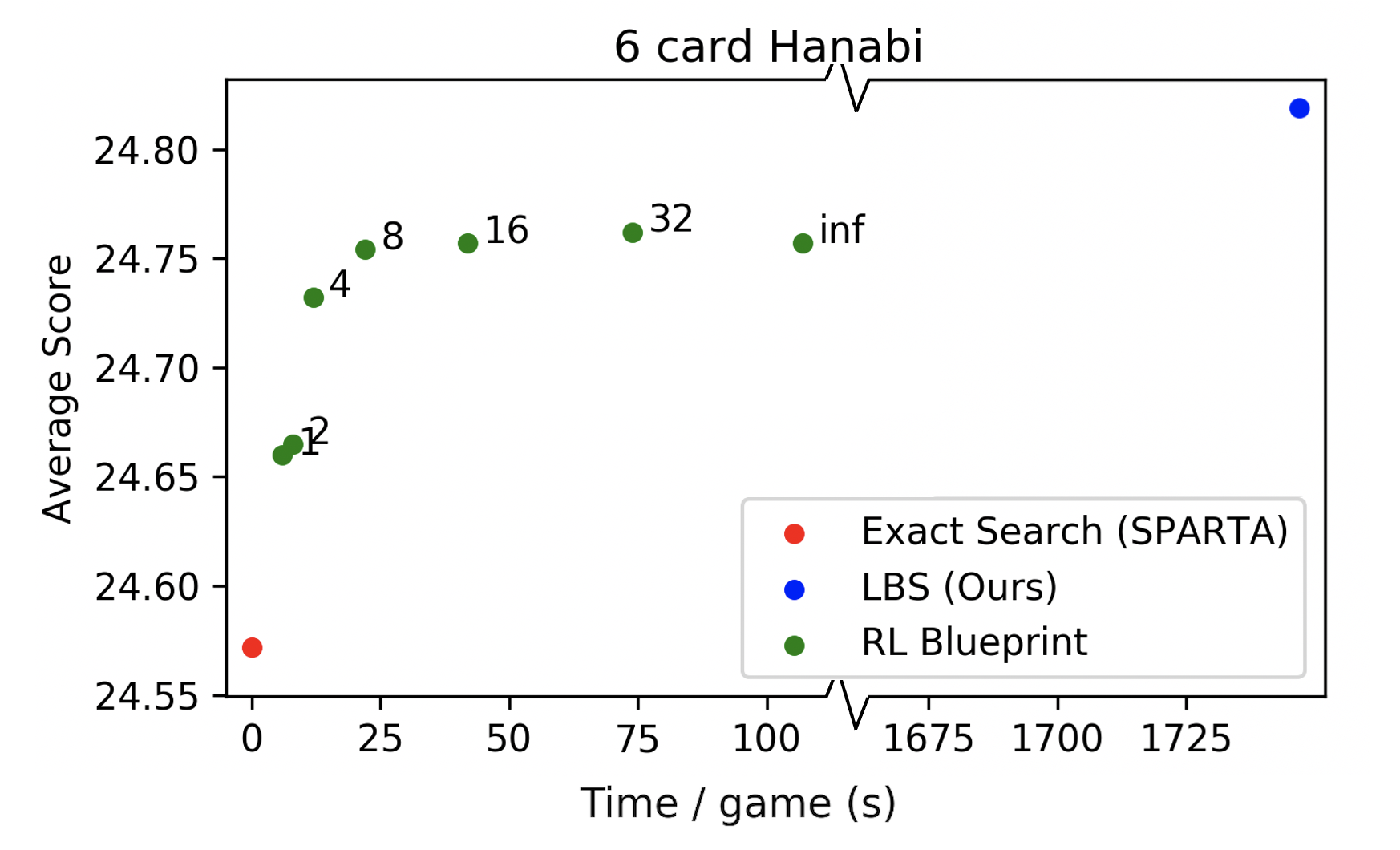}
  \caption{\small Comparison of speed and average score of different policies in 5-card ({\bf top}) and 6-card ({\bf bottom}) Hanabi. The number next to each LBS data point is the rollout depth. LBS with different rollout depths provide a tunable trade-off between speed and performance. LBS provides most of the performance benefits of exact search (SPARTA) at a fraction of the compute cost, and the speedup grows with the size of the belief space (5-card vs 6-card).}
  \label{fig:time_v_score}
\end{figure}

\begin{table*}[t]
\centering
\begin{tabular}{c c c c c c c c}
\toprule
\bf{Variant} & \multicolumn{2}{c}{\bf Blueprint} & \multicolumn{2}{c}{\bf SPARTA}  & \multicolumn{2}{c}{\bf \accmethod} \\
\midrule
{\bf 5-card} & 24.08 $\pm$ 0.01 & ({\it $<$1s}) & 24.52 $\pm$ 0.01 & ({\it 215s}) & 24.48 $\pm$ 0.01 & ({\it 47s})   \\
{\bf 6-card} & 24.57 $\pm$ 0.01 & ({\it $<$1s}) & 24.82 $\pm$ 0.01 &  ({\it 1747s}) & 24.76 $\pm$ 0.01 & ({\it 42s}) \\
{\bf 7-card} & 23.67 $\pm$ 0.02 & ({\it $<$1s}) & \multicolumn{2}{c}{Out-of-Mem Error} & 24.26 $\pm$ 0.01 & ({\it 58s})\\
\bottomrule
\end{tabular}
\caption{\small Result on different Hanabi variants. Each cell contains the mean and standard error of mean (s.e.m.) over 5000 games. It also shows the average time it takes to run a game in the parentheses next to the performance. Note that for the 7-card variant, we reduce the maximum number of hints to 4 in order to make the game harder. All the experiments are executed using 5 CPU cores, 1 Nvidia Quadro GP100 GPU and 64GB of memory. SPARTA fails to run on the 7-card variant due to out-of-memory error since the entire belief space is too large to fit into the allocated memory.}
\label{tab:6-card-result}
% \vspace{-1.5em}
% \end{center}
\end{table*}

\begin{table*}[h]
\begin{center}
\begin{tabular}{l c c c c c | c}
\toprule
\multirow{2}{*}{\bf Method} & \multirow{2}{*}{\bf Depth} & \multirow{2}{*}{\bf Time} & \multicolumn{4}{c}{\bf Blueprint Strength} \\
& & & {\it Weak} & {\it Medium} & {\it Strong} & {\it Best} \\
\midrule
Blueprint & & \it $<$1s & 15.38 $\pm$ 0.05 & 22.99 $\pm$ 0.03  &  24.08 $\pm$ 0.01 & 24.42 $\pm$ 0.01\\
SPARTA & & \it 215s & \textbf{19.53 $\pm$ 0.03} & \textbf{24.16 $\pm$ 0.02} & \textbf{24.52 $\pm$ 0.01} & \textbf{24.66 $\pm$ 0.01} \\
\midrule
\accmethod & $\infty$ & \it 121s & 18.88 $\pm$ 0.03 & 23.95 $\pm$ 0.02 & 24.42 $\pm$ 0.01 & 24.59 $\pm$ 0.01\\
\accmethod & 32 & \it 84s & 19.05 $\pm$ 0.03 & 24.01 $\pm$ 0.02 & 24.45 $\pm$ 0.01 & 24.60 $\pm$ 0.01 \\
\accmethod & 16 & \it 47s & \textbf{19.27 $\pm$ 0.03} & \textbf{24.04 $\pm$ 0.02} & \textbf{24.48 $\pm$ 0.01} & \textbf{24.62 $\pm$ 0.01} \\
\accmethod & 8 & \it 25s & 19.14 $\pm$ 0.03 & 24.03 $\pm$ 0.02 & 24.43 $\pm$ 0.01 & 24.59 $\pm$ 0.01 \\
\accmethod & 4 & \it 14s & 18.75 $\pm$ 0.03 & 23.95 $\pm$ 0.02 & 24.41 $\pm$ 0.01 & 24.55 $\pm$ 0.01 \\
\accmethod & 2 & \it 9s & 18.26 $\pm$ 0.04 & 23.81 $\pm$ 0.02 & 24.35 $\pm$ 0.01 & 24.44 $\pm$ 0.02 \\
\accmethod & 1 & \it 6s & 17.99 $\pm$ 0.04 & 23.69 $\pm$ 0.02 & 24.26 $\pm$ 0.02 & 24.43 $\pm$ 0.02\\
\bottomrule
\end{tabular}
\caption{\small Average scores in 2-player Hanabi with different search variants. {\it Time} column shows the average wall-clock time of each method to play a game. {\it Weak}, {\it Medium} and {\it Strong} blueprints uses 512 hidden units per layer and are trained with RL method after 1 hour, 5 hours, and 20 hours respectively. The {\it Best} blueprint is a larger model with 1024 hidden units each layer trained for 72 hours. Each cell contains the mean and standard error of mean averaged over 5000 games. LBS variants achieve a substantial fraction of the policy improvements of SPARTA over the blueprint at a lower computational cost. Both overall best results (from SPARTA) and the best LBS results are bold for better readability. Note that the {\it Time} column does not reflect the time for the more expensive {\it Best} blueprint.
}
\label{tab:search_result_full}
\end{center}
\end{table*}

Table~\ref{tab:6-card-result} demonstrates the scalability of \accmethod{} compared to SPARTA in environments involving different amount of hidden information. We test them on modified variants of Hanabi where each player holds 6 or 7 cards as well as the official version of 5 cards. In these experiments we use setting mentioned in Section~\ref{sec:bp_training} to train BP with RL and then train belief model until convergence following Section~\ref{sec:belief_training}. It worth noting that it is easier to achieve a higher score with bigger hand size as on average more information will be revealed per hint. In fact, we need to reduce the maximum number of hints from 8 to 4 in the 7 card variant to prevent the RL blueprint policy from being almost perfect.  For comparison, we run SPARTA on our models which can be seen as a expensive upper bound for \accmethod{}. For LBS, we roll out for 16 steps before bootstrapping from the $Q$ values. In the standard Hanabi, LBS delivers 90\% of the performance boost compared to SPARTA while being 4.6$\times$ faster. In the 6-card variant, the SPARTA method runs 8$\times$ slower than it does on standard Hanabi while \accmethod{} runs faster (likely due to shorter games), delivering 76\% of the improvement with 42$\times$ less time. In the 7-card variant, the SPARTA exhausts the 64GB of memory as the storage required for all possible hands and corresponding LSTM states grow exponentially w.r.t. hand size. LBS, on the other hand, scales well and delivers strong performance. Note that the run time of LBS on the 7-card variant is slower partially due to a small change in the implementation to accommodate larger hand.

Figure~\ref{fig:time_v_score} summarizes the speed/performance trade-offs of \accmethod{} compared to running an RL policy directly or using an exact search method (SPARTA). In 5-card Hanabi (left), we start with a blueprint policy that achieves an average score of 22.99 after 5 hours of training, and apply \accmethod{} with different rollout lengths as well as an exact search. \accmethod{} achieves most of the score improvement of exact search while being much faster to execute. Changing the rollout depth allows for tuning speed vs. performance. As we move to a larger game (6-card Hanabi), SPARTA becomes 10$\times$ slower while \accmethod{} runs at approximately the same speed and continues to provide a similar fraction of the performance improvement.

As mentioned in Section~\ref{sec:method}, applying LBS independently on multiple players will not lead to better performance due to the inaccurate beliefs. To verify this, we run LBS on both players in standard 5-card Hanabi. The result is 24.34 $\pm$ 0.02, which is noticeably worse than the 24.48 $\pm$ 0.01 achieved by single agent LBS.

A more comprehensive analysis of these trade-offs is shown in Table~\ref{tab:search_result_full}. We train the BP with RL for a total of 20 hours and use 3 snapshots to demonstrate the performance of \accmethod{} given BPs of different strength.  {\it Weak}, {\it Medium} and {\it Strong} policies are snapshots after 1 hour, 5 hours and 20 hours respectively. The network uses 512 units for both LSTM and feed-forward layers. The belief models are trained to convergence for each BP. All search methods run 10K rollouts per step. The best performing variant is \accmethod{} with 16-step rollouts, delivering 91\% of the improvement on average comparing to the exact search while being 4.6$\times$ faster. Even the cheapest method, \accmethod{} with 1-step rollouts, returns a decent improvement of 55\% on average and 35.8$\times$ speedup. It also worth noting that LBS performs strongly even on the weakest policy, demonstrating high versatility of the method.

To further test the limit of self-play in Hanabi, we train a larger model with 1024 units per layer for an extend period of 72 hours and apply both LBS and SPARTA on top. The model is dubbed {\it Best} in Table~\ref{tab:search_result_full}. The BP itself gets 24.42 and SPARTA further boosts the performance to 24.66, the highest self-play score ever reported in 2 player Hanabi. Remarkably, LBS achieves 24.62, delivering 83\% fo the benefit of SPARTA while taking less than $1/4$ of the time (69s for LBS and 364s for SPARTA).

We note that the more expensive \accmethod{} that rolls out until the end of the games (LBS-$\infty$) is not the best performing one. It consistently under-performs some of the \accmethod-k variants by a small, in many cases significant, amount. 
We hypothesis that under \accmethod-$\infty$ it is more likely for the agents to reach a state that is under-explored during training. Therefore the approximate belief will be less accurate and the estimate Q-value be wrong. The \accmethod-k methods where the Q-value after k steps is used as a bootstrap may naturally avoid those situations since the BP may also have low Q values for under-explored states. One piece of evidence for this theory is that in \accmethod-$\infty$, 0.1\% of the games end up with a completely failed belief prediction and have to revert back to BP while the same only happens to 0.0083\% of the games for \accmethod-k. 

Clearly, this is a potential problem: part of the reason search works well is that it can discover moves that were underestimated  and, consequently, under-explored  by the BP. The brittleness of the learned belief, in combination with the difference in overall belief quality (Fig~\ref{fig:belief-perf}), help explain the difference in performance between \accmethod{} and exact methods like SPARTA.

\subsection{Belief Quality}

\begin{figure}[t]%{0.45\textwidth}
\centering
\includegraphics[width=0.4\textwidth]{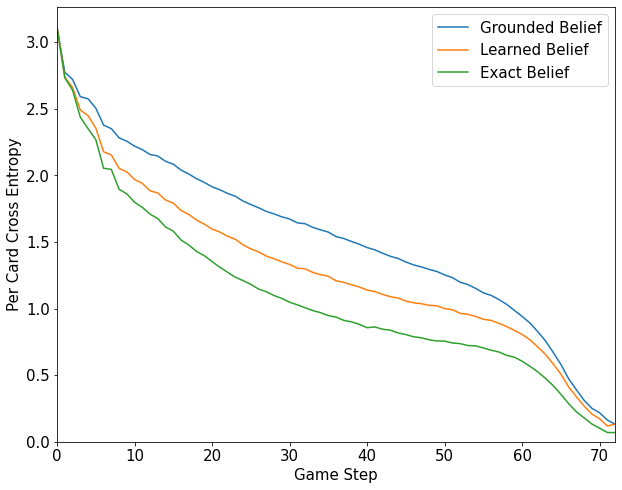}
\caption{\small Per-card cross entropy with the true hand for different beliefs in games played by BP.}
\label{fig:belief-perf}
% \vspace{-2em}
\end{figure}

We examine the quality of the learned belief model by looking at its cross entropy loss for predicting hidden card values.   We  compare  against  two  benchmarks: the exact beliefs marginalized over each card, and an auto-regressive belief based only on grounded information.  The grounded belief predicts a distribution over card values proportional to remaining card counts, for all card values consistent with past hints. 

We generate 1000 games with the strong policy and compute the two benchmarks together with the loss of the learned belief (Eq.~\ref{eq:belief-loss}) from a fully trained belief model. Figure~\ref{fig:belief-perf} shows how these 3 values change over the course of games. We see that our learned belief model performs considerably better than the grounded belief. There is still a gap between learned belief and exact belief, especially in the later stage of the game. More powerful models such as transformers~\citep{transformer} may further improve the belief learning but we leave it for future work.

\begin{figure}[t]
\includegraphics[width=0.45\textwidth]{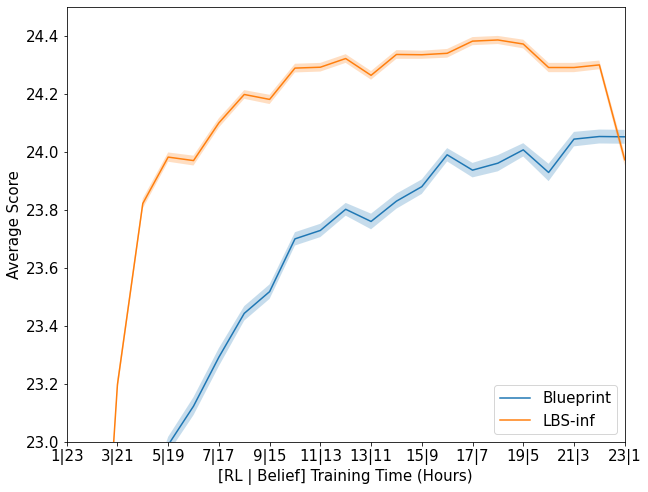}
\caption{\small Result of fixed budget (24 hours) at training time. The ticks ``$a|b$''on x-axis means $a$ hours to train BP and $b$ hours to train belief model.} 
\label{fig:fix-budget-train}
% \end{figure}
% \begin{figure}[h]
\includegraphics[width=0.44\textwidth]{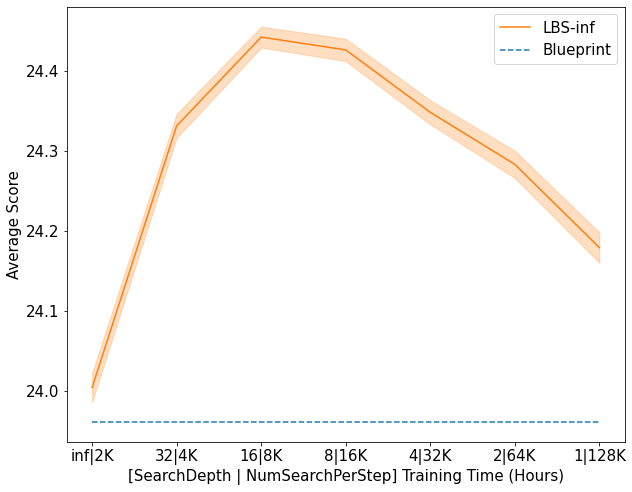}
\caption{\small Result of fixed budget at test time. The ticks ``$a|b$'' on x-axis means run search for $a$ steps before bootstrapping with Q function and run $b$ search per move. Each data point on both figures is evaluated on 5000 games and the shaded area is the standard error of mean.}
\label{fig:fix-budget-search}
\end{figure}

\subsection{Fixed Budget Training \& Testing}

Since one of the motivations for \accmethod~is speed, it would be interesting to know how we could allocate resources at training and test time to maximize performance given a fixed computational budget. For fixed training budget, we train the BP for $l$ hours and then train belief model for $24-l$ hours. We evaluate these combinations with \accmethod-$\infty$ as well as BPs themselves. As shown in Figure~\ref{fig:fix-budget-train}, with longer RL training, the BP improves monotonically, but the final performance suffers due to a poor belief model. The best combination is $\sim$18 hours for RL and $\sim$6 hours for belief learning.

We then take the models from the best combination to study how to allocate compute between the number of rollouts and rollout depth. We start with \accmethod-1 and 128K search per step, and then halve the number of searches as we double the search depth. 
The result is shown in Figure~\ref{fig:fix-budget-search}. If we compare the results here with those from Table~\ref{tab:search_result_full}, we see that although \accmethod-16 still has the best performance, the relative strength between \accmethod-32 and \accmethod-8 flips, indicating that the trade-off may still matter in some cases.

\section{Conclusion}
\label{sec:conclusion}
We presented Learned Belief Search, a novel search algorithms for POMDPs that can be used to improve upon the performance of a blueprint policy at test time whenever a simulator of the environment is available. We also presented extensions of LBS that make it applicable to fully cooperative, partially observable multi-agent settings. 
The heart of LBS is an auto-regressive model that can be used to generate samples from an approximate belief for any given AOH. 

While LBS achieves strong performance on the benchmark problem Hanabi, our work also clearly points to a number of future directions. For a start, the search process can bring the belief model to under-explored regions of the state space. This could be addressed by retraining the belief model on the data generated from LBS. 

Another interesting direction for future work is to amortize the search process, \emph{e.g.} by integrating it into the training process, and to extend LBS to multiplayer and multi-step search.
To enable these directions, and many others, we plan to open-source all of the code for LBS. 

\bibliography{references}

\begin{thebibliography}{29}
\providecommand{\natexlab}[1]{#1}
\providecommand{\url}[1]{\texttt{#1}}
\expandafter\ifx\csname urlstyle\endcsname\relax
  \providecommand{\doi}[1]{doi: #1}\else
  \providecommand{\doi}{doi: \begingroup \urlstyle{rm}\Url}\fi

\bibitem[Bard et~al.(2020)Bard, Foerster, Chandar, Burch, Lanctot, Song,
  Parisotto, Dumoulin, Moitra, Hughes, Dunning, Mourad, Larochelle, Bellemare,
  and Bowling]{bard2019hanabi}
Bard, N., Foerster, J.~N., Chandar, S., Burch, N., Lanctot, M., Song, H.~F.,
  Parisotto, E., Dumoulin, V., Moitra, S., Hughes, E., Dunning, I., Mourad, S.,
  Larochelle, H., Bellemare, M.~G., and Bowling, M.
\newblock The hanabi challenge: A new frontier for ai research.
\newblock \emph{Artificial Intelligence}, 280:\penalty0 103216, 2020.
\newblock ISSN 0004-3702.

\bibitem[Bertsekas \& Castanon(1999)Bertsekas and
  Castanon]{bertsekas1999rollout}
Bertsekas, D.~P. and Castanon, D.~A.
\newblock Rollout algorithms for stochastic scheduling problems.
\newblock \emph{Journal of Heuristics}, 5\penalty0 (1):\penalty0 89--108, 1999.

\bibitem[Brown \& Sandholm(2017)Brown and Sandholm]{brown2017superhuman}
Brown, N. and Sandholm, T.
\newblock Superhuman {A}{I} for heads-up no-limit poker: Libratus beats top
  professionals.
\newblock \emph{Science}, pp.\  eaao1733, 2017.

\bibitem[Brown \& Sandholm(2019)Brown and Sandholm]{brown2019superhuman}
Brown, N. and Sandholm, T.
\newblock Superhuman {A}{I} for multiplayer poker.
\newblock \emph{Science}, pp.\  eaay2400, 2019.

\bibitem[Campbell et~al.(2002)Campbell, Hoane~Jr, and Hsu]{campbell2002deep}
Campbell, M., Hoane~Jr, A.~J., and Hsu, F.-h.
\newblock Deep {B}lue.
\newblock \emph{Artificial intelligence}, 134\penalty0 (1-2):\penalty0 57--83,
  2002.

\bibitem[Canaan et~al.(2019)Canaan, Togelius, Nealen, and
  Menzel]{canaan2019diverse}
Canaan, R., Togelius, J., Nealen, A., and Menzel, S.
\newblock Diverse agents for ad-hoc cooperation in hanabi.
\newblock In \emph{IEEE Conference on Games 2019, CoG 2019}, IEEE Conference on
  Computatonal Intelligence and Games, CIG. IEEE Computer Society, August 2019.
\newblock \doi{10.1109/CIG.2019.8847944}.
\newblock 2019 IEEE Conference on Games, CoG 2019 ; Conference date: 20-08-2019
  Through 23-08-2019.

\bibitem[Foerster et~al.(2019)Foerster, Song, Hughes, Burch, Dunning, Whiteson,
  Botvinick, and Bowling]{foerster2019bayesian}
Foerster, J., Song, F., Hughes, E., Burch, N., Dunning, I., Whiteson, S.,
  Botvinick, M., and Bowling, M.
\newblock Bayesian action decoder for deep multi-agent reinforcement learning.
\newblock In \emph{International Conference on Machine Learning}, pp.\
  1942--1951, 2019.

\bibitem[Hausknecht \& Stone(2015)Hausknecht and Stone]{hausknecht2015deep}
Hausknecht, M. and Stone, P.
\newblock Deep recurrent q-learning for partially observable mdps.
\newblock In \emph{2015 AAAI Fall Symposium Series}, 2015.

\bibitem[Hor{\'a}k \& Bo{\v{s}}ansk{\`y}(2019)Hor{\'a}k and
  Bo{\v{s}}ansk{\`y}]{horak2019solving}
Hor{\'a}k, K. and Bo{\v{s}}ansk{\`y}, B.
\newblock Solving partially observable stochastic games with public
  observations.
\newblock In \emph{Proceedings of the AAAI Conference on Artificial
  Intelligence}, volume~33, pp.\  2029--2036, 2019.

\bibitem[Hu \& Foerster(2020)Hu and Foerster]{Hu2020Simplified}
Hu, H. and Foerster, J.~N.
\newblock Simplified action decoder for deep multi-agent reinforcement
  learning.
\newblock In \emph{International Conference on Learning Representations}, 2020.
\newblock URL \url{https://openreview.net/forum?id=B1xm3RVtwB}.

\bibitem[Hu et~al.(2020)Hu, Peysakhovich, Lerer, and Foerster]{hu2020other}
Hu, H., Peysakhovich, A., Lerer, A., and Foerster, J.
\newblock \textquotedblleft other-play\textquotedblright for zero-shot
  coordination.
\newblock In \emph{Proceedings of Machine Learning and Systems 2020}, pp.\
  9396--9407. 2020.

\bibitem[Hu et~al.(2021)Hu, Lerer, Cui, Pineda, Wu, Brown, and Foerster]{obl}
Hu, H., Lerer, A., Cui, B., Pineda, L., Wu, D., Brown, N., and Foerster, J.~N.
\newblock Off-belief learning.
\newblock \emph{CoRR}, abs/2103.04000, 2021.
\newblock URL \url{https://arxiv.org/abs/2103.04000}.

\bibitem[Kova{\v{r}}{\'\i}k et~al.(2019)Kova{\v{r}}{\'\i}k, Schmid, Burch,
  Bowling, and Lis{\`y}]{kovavrik2019rethinking}
Kova{\v{r}}{\'\i}k, V., Schmid, M., Burch, N., Bowling, M., and Lis{\`y}, V.
\newblock Rethinking formal models of partially observable multiagent decision
  making.
\newblock \emph{arXiv preprint arXiv:1906.11110}, 2019.

\bibitem[Lerer et~al.(2020)Lerer, Hu, Foerster, and Brown]{lerer2020improving}
Lerer, A., Hu, H., Foerster, J.~N., and Brown, N.
\newblock Improving policies via search in cooperative partially observable
  games.
\newblock In \emph{AAAI}, pp.\  7187--7194, 2020.

\bibitem[Morav{\v{c}}{\'\i}k et~al.(2017)Morav{\v{c}}{\'\i}k, Schmid, Burch,
  Lis{\`y}, Morrill, Bard, Davis, Waugh, Johanson, and
  Bowling]{moravvcik2017deepstack}
Morav{\v{c}}{\'\i}k, M., Schmid, M., Burch, N., Lis{\`y}, V., Morrill, D.,
  Bard, N., Davis, T., Waugh, K., Johanson, M., and Bowling, M.
\newblock Deepstack: Expert-level artificial intelligence in heads-up no-limit
  poker.
\newblock \emph{Science}, 356\penalty0 (6337):\penalty0 508--513, 2017.

\bibitem[O'Dwyer(2019)]{quuxplusone}
O'Dwyer, A.
\newblock Hanabi.
\newblock \url{https://github.com/Quuxplusone/Hanabi}, 2019.

\bibitem[Ross et~al.(2008)Ross, Pineau, Paquet, and Chaib-Draa]{ross2008online}
Ross, S., Pineau, J., Paquet, S., and Chaib-Draa, B.
\newblock Online planning algorithms for pomdps.
\newblock \emph{Journal of Artificial Intelligence Research}, 32:\penalty0
  663--704, 2008.

\bibitem[Roy et~al.(2005)Roy, Gordon, and Thrun]{roy2005finding}
Roy, N., Gordon, G., and Thrun, S.
\newblock Finding approximate pomdp solutions through belief compression.
\newblock \emph{Journal of artificial intelligence research}, 23:\penalty0
  1--40, 2005.

\bibitem[Schrittwieser et~al.(2019)Schrittwieser, Antonoglou, Hubert, Simonyan,
  Sifre, Schmitt, Guez, Lockhart, Hassabis, Graepel,
  et~al.]{schrittwieser2019mastering}
Schrittwieser, J., Antonoglou, I., Hubert, T., Simonyan, K., Sifre, L.,
  Schmitt, S., Guez, A., Lockhart, E., Hassabis, D., Graepel, T., et~al.
\newblock Mastering atari, go, chess and shogi by planning with a learned
  model.
\newblock \emph{arXiv preprint arXiv:1911.08265}, 2019.

\bibitem[Silver \& Veness(2010)Silver and Veness]{silver2010monte}
Silver, D. and Veness, J.
\newblock Monte-carlo planning in large pomdps.
\newblock In \emph{Advances in neural information processing systems}, pp.\
  2164--2172, 2010.

\bibitem[Silver et~al.(2016)Silver, Huang, Maddison, Guez, Sifre, Van
  Den~Driessche, Schrittwieser, Antonoglou, Panneershelvam, Lanctot,
  et~al.]{silver2016mastering}
Silver, D., Huang, A., Maddison, C.~J., Guez, A., Sifre, L., Van Den~Driessche,
  G., Schrittwieser, J., Antonoglou, I., Panneershelvam, V., Lanctot, M.,
  et~al.
\newblock Mastering the game of go with deep neural networks and tree search.
\newblock \emph{Nature}, 529\penalty0 (7587):\penalty0 484, 2016.

\bibitem[Silver et~al.(2017)Silver, Schrittwieser, Simonyan, Antonoglou, Huang,
  Guez, Hubert, Baker, Lai, Bolton, et~al.]{silver2017mastering}
Silver, D., Schrittwieser, J., Simonyan, K., Antonoglou, I., Huang, A., Guez,
  A., Hubert, T., Baker, L., Lai, M., Bolton, A., et~al.
\newblock Mastering the game of go without human knowledge.
\newblock \emph{Nature}, 550\penalty0 (7676):\penalty0 354, 2017.

\bibitem[Silver et~al.(2018)Silver, Hubert, Schrittwieser, Antonoglou, Lai,
  Guez, Lanctot, Sifre, Kumaran, Graepel, et~al.]{silver2018general}
Silver, D., Hubert, T., Schrittwieser, J., Antonoglou, I., Lai, M., Guez, A.,
  Lanctot, M., Sifre, L., Kumaran, D., Graepel, T., et~al.
\newblock A general reinforcement learning algorithm that masters chess, shogi,
  and go through self-play.
\newblock \emph{Science}, 362\penalty0 (6419):\penalty0 1140--1144, 2018.

\bibitem[Tesauro(1994)]{tesauro1994td}
Tesauro, G.
\newblock T{D}-{G}ammon, a self-teaching backgammon program, achieves
  master-level play.
\newblock \emph{Neural computation}, 6\penalty0 (2):\penalty0 215--219, 1994.

\bibitem[Tian et~al.(2020)Tian, Gong, and Jiang]{tian2020joint}
Tian, Y., Gong, Q., and Jiang, T.
\newblock Joint policy search for multi-agent collaboration with imperfect
  information.
\newblock \emph{arXiv preprint arXiv:2008.06495}, 2020.

\bibitem[Vaswani et~al.(2017)Vaswani, Shazeer, Parmar, Uszkoreit, Jones, Gomez,
  Kaiser, and Polosukhin]{transformer}
Vaswani, A., Shazeer, N., Parmar, N., Uszkoreit, J., Jones, L., Gomez, A.~N.,
  Kaiser, L.~u., and Polosukhin, I.
\newblock Attention is all you need.
\newblock In Guyon, I., Luxburg, U.~V., Bengio, S., Wallach, H., Fergus, R.,
  Vishwanathan, S., and Garnett, R. (eds.), \emph{Advances in Neural
  Information Processing Systems 30}, pp.\  5998--6008. Curran Associates,
  Inc., 2017.
\newblock URL
  \url{http://papers.nips.cc/paper/7181-attention-is-all-you-need.pdf}.

\bibitem[Walton-Rivers et~al.(2017)Walton-Rivers, Williams, Bartle,
  Perez-Liebana, and Lucas]{walton2017evaluating}
Walton-Rivers, J., Williams, P.~R., Bartle, R., Perez-Liebana, D., and Lucas,
  S.~M.
\newblock Evaluating and modelling hanabi-playing agents.
\newblock In \emph{IEEE Congress on Evolutionary Computation (CEC)}, pp.\
  1382--1389, 2017.

\bibitem[Wang et~al.(2016)Wang, Schaul, Hessel, Hasselt, Lanctot, and
  Freitas]{dueling-dqn}
Wang, Z., Schaul, T., Hessel, M., Hasselt, H., Lanctot, M., and Freitas, N.
\newblock Dueling network architectures for deep reinforcement learning.
\newblock volume~48 of \emph{Proceedings of Machine Learning Research}, pp.\
  1995--2003, New York, New York, USA, 20--22 Jun 2016. PMLR.
\newblock URL \url{http://proceedings.mlr.press/v48/wangf16.html}.

\bibitem[Wu(2018)]{FireFlower}
Wu, D.
\newblock A rewrite of hanabi-bot in scala.
\newblock \url{https://github.com/lightvector/fireflower}, 2018.

\end{thebibliography}
\bibliographystyle{icml2021}

% \clearpage
% \newpage
% \appendix
% \input{appendix/belief}
% \input{appendix/full_lbs_result}

\end{document}

\twocolumn[
\icmltitle{Supplementary Material}



\icmlsetsymbol{equal}{*}

% \begin{icmlauthorlist}
% \icmlauthor{Aeiau Zzzz}{equal,to}
% \icmlauthor{Bauiu C.~Yyyy}{equal,to,goo}
% \icmlauthor{Cieua Vvvvv}{goo}
% \icmlauthor{Iaesut Saoeu}{ed}
% \icmlauthor{Fiuea Rrrr}{to}
% \icmlauthor{Tateu H.~Yasehe}{ed,to,goo}
% \icmlauthor{Aaoeu Iasoh}{goo}
% \icmlauthor{Buiui Eueu}{ed}
% \icmlauthor{Aeuia Zzzz}{ed}
% \icmlauthor{Bieea C.~Yyyy}{to,goo}
% \icmlauthor{Teoau Xxxx}{ed}
% \icmlauthor{Eee Pppp}{ed}
% \end{icmlauthorlist}

% \icmlaffiliation{to}{Department of Computation, University of Torontoland, Torontoland, Canada}
% \icmlaffiliation{goo}{Googol ShallowMind, New London, Michigan, USA}
% \icmlaffiliation{ed}{School of Computation, University of Edenborrow, Edenborrow, United Kingdom}

% \icmlcorrespondingauthor{Cieua Vvvvv}{c.vvvvv@googol.com}
% \icmlcorrespondingauthor{Eee Pppp}{ep@eden.co.uk}

% % You may provide any keywords that you
% % find helpful for describing your paper; these are used to populate
% % the "keywords" metadata in the PDF but will not be shown in the document

\vskip 0.3in
\appendix

% \section{Appendix}

]
\appendix
\onecolumn

\section{Additional Details on Experimental Setup}
\label{sec:detail-experimental-setup}
As mentioned in the main text, we base our experiments on the existing implementations of Other-Play and SPARTA for training blueprint and running search respectively and therefore inherit many of their hyper-parameters. For completeness we include a detailed documentation of the experimental setup here.

The blueprint policy training is set up in a distributed fashion. On the simulation side, 6400 Hanabi simulators run in parallel and the observations produced by those simulators are batched together for efficient neural network computation on GPU. Due to this highly efficient setup, a single machine with 80 CPU cores and 2 GPUs is sufficient to run entire simulation workload while training runs on an additional third GPU. The simulated trajectories are collected into a prioritized replay buffer~\cite{prioritized-replay}. The priority of each trajectory is computed as $\xi = 0.9 * \max_{i} \xi_{i} + 0.1 * \bar{\xi}$ ~\citep{kapturowski2018recurrent} where $\xi_{i}$ is the per step absolute TD error. During training, we sample mini-batches from the replay buffer to update the policy. The simulation policy is synced with the policy being trained every 10 gradient steps. 
The exploration is handled by epsilon greedy method. At the beginning of each simulated game, each player samples an epsilon independently from $\epsilon_{i}=\alpha^{1 + \beta * \frac{i}{N-1}}$ 
where $\alpha = 0.1$, $\beta=7$ and $N=80$. We follow~\cite{hu2020other} to use the Other-Play objective. We also use the auxiliary task  predicting the playable/discardable/unknown properties of the hidden cards as introduced in~\cite{Hu2020Simplified}. The weight of the auxiliary loss is set to 0.25. The rest of the hyper-parameters can be found in Table~\ref{tab:hparam-rl}.

% \begin{multicols}{3} {
\begin{table*}[]
    \centering
    \begin{tabular}{l l}
    \toprule
    Hyper-parameters     &  Value \\
    \midrule
    \texttt{\# replay buffer related}     &  \\
    \texttt{burn\_in\_frames} & \texttt{5000} \\
    \texttt{replay\_buffer\_size} & \texttt{$2^{17}$} \\
    \texttt{priority\_exponent} & \texttt{0.9} \\
    \texttt{priority\_weight} & \texttt{0.6} \\
    \texttt{max\_trajectory\_length} & \texttt{80} \\
    \midrule
    \texttt{\# optimization} & \\
    \texttt{optimizer} & \texttt{Adam~\citep{kingma2014adam}} \\
    \texttt{lr} & \texttt{6.25e-05} \\
    \texttt{eps} & \texttt{1.5e-05} \\
    \texttt{grad\_clip} & \texttt{5} \\
    \texttt{batchsize} & \texttt{64} \\
    \midrule
    \texttt{\# Q learning} & \\
    \texttt{n\_step} & \texttt{3} \\
    \texttt{discount\_factor} & \texttt{0.999} \\
    \texttt{target\_network\_sync\_interval} & \texttt{2500} \\
    \bottomrule
    \end{tabular}
    \caption{\small Hyper-Parameters for Reinforcement Learning}
    \label{tab:hparam-rl}
\end{table*}
% }

The belief model is trained under a similar setting as blueprint policy with 3 major modifications. Firstly we load a pretrained policy for running simulations and keep it fixed. Secondly the trajectories are drawn uniformly from the replay buffer with no priorities. Finally the model is trained with the auto-regressive maximum likelihood loss:
\begin{equation}
\mathcal{L}(c_{1:n} | \tau) = -\sum_{i=1}^{n} \log p(c_i | \tau, c_{1:i-1}),
\label{eq:belief-loss}
\end{equation}
with supervised learning. The belief model is trained with Adam optimizer with learning rate $2.5\times10^{-4}$ and \texttt{eps} = $1.5\times 10^{-5}$.

The search experiments are benchmarked under the resource constraint of 5 CPU cores, 1 Nvidia Quadro GP100 GPU and 64GB of memory. We perform 10K searches at each step except for experiments for Figure~\ref{fig:fix-budget-search}. Searches run parallel over multiple CPU cores while neural network computations are batched across searches for maximum efficiency. The search will only change the move if the expected value of the action chosen by search is $\delta=0.05$ higher than that chosen by the BP. A UCB-like pruning method is used to reduce the number of samples required.

\bibliography{references}
\bibliographystyle{icml2021}